\pdfoutput=1
\documentclass[11pt]{article}
\usepackage[final]{acl}
\usepackage{times}
\usepackage{latexsym}
\usepackage{float}
\usepackage{listings}
\usepackage{amsmath}
\usepackage[T1]{fontenc}
\usepackage[utf8]{inputenc}
\usepackage{microtype}
\usepackage{inconsolata}
\usepackage{graphicx}
\title{1-800-SHARED-TASKS @ NLU of Devanagari Script Languages: Detection of Language, Hate Speech, and Targets using LLMs}
\author{Jebish Purbey \\
  Pulchowk Campus, IoE \\
  \texttt{jebishpurbey@gmail.com} \\
  \And Siddartha Pullakhandam \\
  University of Wisconsin \\
  \texttt{pullakh2@uwm.edu} \\ 
  \And Kanwal Mehreen * \\
  Traversaal.ai \\
  \texttt{kanwal@traversaal.ai} \\ \\
  \AND Muhammad Arham * \\
  NUST/Traversaal.ai \\
  \texttt{arhamm40182@gmail.com} \\ \\
  \And Drishti Sharma * \\
  Cohere For AI Community\\
  \texttt{drishtishrma@gmail.com} \\ \\
  \And Ashay Srivastava \\
  University of Maryland \\
  \texttt{ashays06@umd.edu} \\
  \AND Ram Mohan Rao Kadiyala  \\
  University of Maryland, College Park \\
  \texttt{rkadiyal@umd.edu} \\ }
\begin{document}
\maketitle
\begin{abstract}
This paper presents a detailed system description of our entry for the CHiPSAL 2025 shared task, focusing on language detection, hate speech identification, and target detection in Devanagari script languages. We experimented with a combination of large language models and their ensembles, including MuRIL, IndicBERT, and Gemma-2, and leveraged unique techniques like focal loss to address challenges in the natural understanding of Devanagari languages, such as multilingual processing and class imbalance. Our approach achieved competitive results across all tasks: F1 of 0.9980, 0.7652, and 0.6804 for Sub-tasks A, B, and C respectively. This work provides insights into the effectiveness of transformer models in tasks with domain-specific and linguistic challenges, as well as areas for potential improvement in future iterations.
\end{abstract}
\makeatletter
\def\blfootnote{\gdef\@thefnmark{}\@footnotetext}
\makeatother
\blfootnote{* equal contribution}
\section{Introduction}
Large language models (LLMs) have revolutionized natural language processing (NLP) yet South Asian languages remain largely underrepresented within these advancements despite being home to over 700 languages, 25 major scripts, and approximately 1.97 billion people. Addressing these gaps, this paper focuses on three critical NLP tasks of CHiPSAL 2025 \cite{sarves2025chipsal} in Devanagari-scripted languages: 5-way classification of the text based on the language of the text (Sub-task A), Binary classification for detecting hate speech in the text (Sub-task B), and 3-way classification for detecting target of hate speech in a text (Sub-task C) \cite{thapa2025nludevanagari}. Our system leverages the multilingual capabilities of open-source LLMs namely IndicBERT V2 \cite{doddapaneni-etal-2023-towards},  MuRIL \cite{khanuja2021murilmultilingualrepresentationsindian}, and Gemma-2 \cite{gemmateam2024gemmaopenmodelsbased} and their ensembles for natural language understanding of Devanagari script languages. Our work contributes to advancing language technology in South Asia, aiming for inclusivity and deeper understanding across diverse linguistic landscapes.
\section{Dataset \& Task}
The goal of Sub-task A is to determine the language of the given Devanagari script among the 5 languages to address the critical need for accurate multilingual identification. The dataset consists of text in Nepali \cite{thapa2023nehate,rauniyar2023multi}, Marathi \cite{kulkarni2021l3cubemahasent}, Sanskrit \cite{aralikatte2021itihasa}, Bhojpuri \cite{ojha2019english}, and Hindi \cite{jafri2024chunav,jafri2023uncovering}. For Sub-task B, the goal is to determine if the text contains hate speech or not. The dataset consists of social media text (tweets) in Hindi and Nepali languages. Sub-task C follows Sub-task B, where the goal is to identify the targets of hate speech among "individual", "organization", or "community". Similar to Sub-task B, the dataset for Sub-task C is in Hindi and Nepali languages. The distribution of labels for the three datasets can be seen in table \ref{tab:datasetA}, \ref{tab:datasetB}, and \ref{tab:datasetC} respectively.
\begin{table}[h]
\centering
\begin{tabular}{|l|c|c|c|c|}
\hline
\textbf{Class}    & \textbf{Train} & \textbf{Dev} & \textbf{Test} \\ \hline
\textbf{Nepali}   & 12544          & 2688         & 2688          \\
\textbf{Marathi}  & 11034          & 2364         & 2365          \\
\textbf{Sanskrit} & 10996          & 2356         & 2356          \\
\textbf{Bhojpuri} & 10184          & 2182         & 2183          \\
\textbf{Hindi}    & 7664           & 1643         & 1642          \\ \hline
\textbf{Total}    & 52422          & 11233        & 11234         \\ \hline
\end{tabular}
\caption{Class distribution for Subtask A}
\label{tab:datasetA}
\end{table}

\begin{table}[t]
\centering
\begin{tabular}{|l|c|c|c|c|}
\hline
\textbf{Class}    & \textbf{Train} & \textbf{Dev} & \textbf{Test} \\ \hline
\textbf{Non-hate} & 16805          & 3602         & 3601          \\
\textbf{Hate}     & 2214           & 474          & 475           \\ \hline
\textbf{Total}    & 19019          & 4076         & 4076          \\ \hline
\end{tabular}
\caption{Class distribution for the Subtask B}
\label{tab:datasetB}
\end{table}

\begin{figure*}[t]
    \centering
  \includegraphics[scale=0.47]{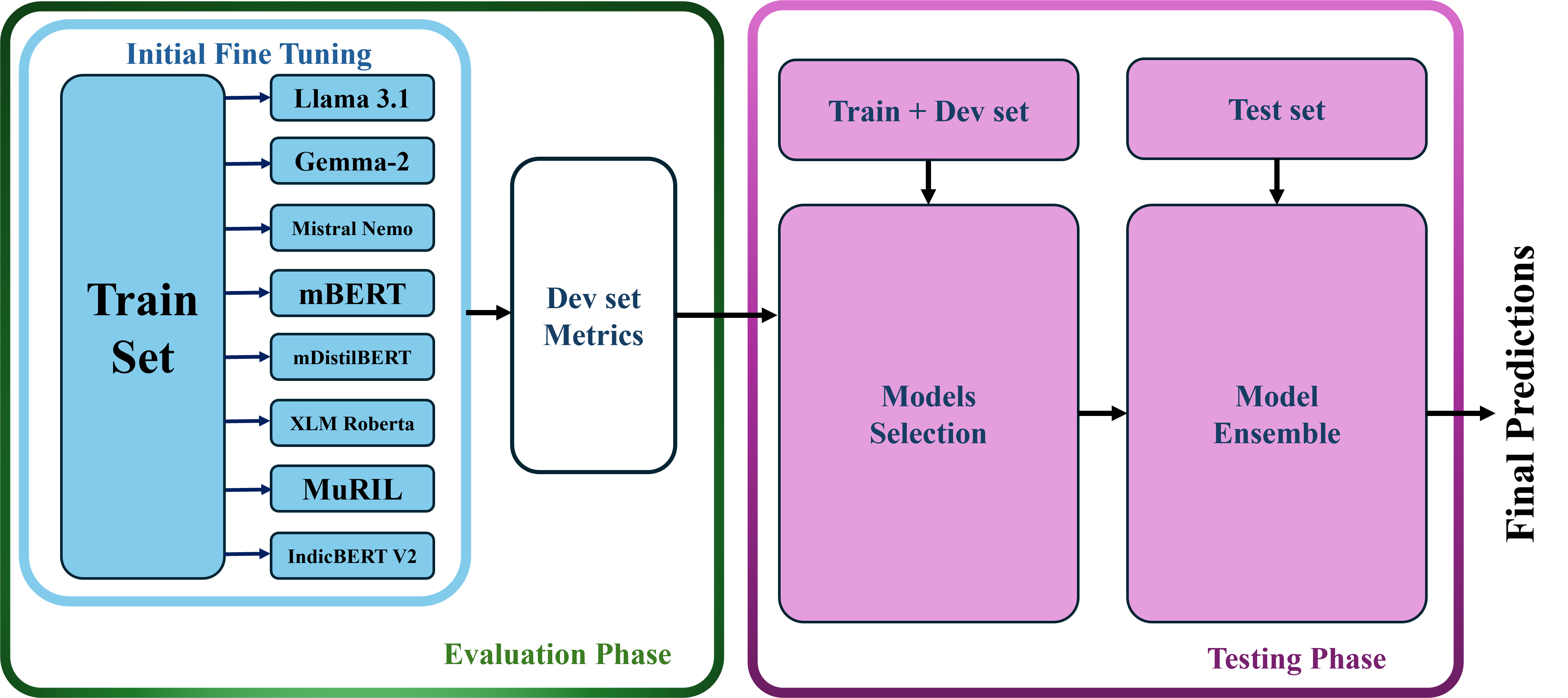}
  \caption{System design workflow. The development set is initially used to select the best-performing models, which are then retrained on the combined train and development set. Selected models are ensembled to generate final predictions on the test set.}
  \label{fig:system}
\end{figure*}

\begin{table}[t]
\centering
\begin{tabular}{|l|c|c|c|c|}
\hline
\textbf{Class}                           & \textbf{Train} & \textbf{Dev} & \textbf{Test} \\ \hline
\textbf{Individual}                      & 1074           & 230          & 230           \\
\textbf{Organization}                    & 856            & 183          & 184           \\
\textbf{Community}                       & 284            & 61           & 61            \\ \hline
\textbf{Total}                           & 2214           & 474          & 475           \\ \hline
\end{tabular}
\caption{Class distribution for the Subtask C}
\label{tab:datasetC}
\end{table}

\section{Methodology}

The common approach to all three Sub-tasks was to fine-tune a multitude of multilingual models in the train set and use the dev set to select the best few models during the Evaluation phase. The selected best models were then fine-tuned again on both the train and dev sets and their ensemble, by majority voting, was used for the final prediction of the test set during the Testing phase as shown in Figure \ref{fig:system}. The models fine-tuned under this approach include decoder-only models such as Gemma-2 9B, Llama 3.1 8B \cite{dubey2024llama3herdmodels}, and Mistral Nemo Base 12B \cite{mistral_nemo}, and BERT \cite{devlin2019bertpretrainingdeepbidirectional} based models such as IndicBERT V2, MuRIL, XLM Roberta \cite{DBLP:journals/corr/abs-1911-02116}, mDistilBERT \cite{Sanh2019DistilBERTAD} and mBERT \cite{DBLP:journals/corr/abs-1810-04805}.
For decoder-only models, each Sub-task was formulated as a text-generation task where each model was asked to generate only one option among the given choices. For BERT-based models, each Sub-task was formulated as a multilabel classification task by adding a classification head to the model.\\
For Sub-task A, each decoder-only models were trained for 1 epoch with a learning rate of 2e-4. The BERT-based models were trained for 5 epochs with a learning rate of 4e-5 with weighted cross-entropy loss. For Sub-task B, decoder-only models were trained for 2-4 epochs with a learning rate of 2e-4. The BERT-based models were trained for 5 epochs with a learning rate of 4e-5.\\
To handle the class imbalance in sub-task B, focal loss \cite{lin2018focallossdenseobject} was used for BERT-based models. Focal loss modifies cross-entropy by reducing the relative loss for well-classified examples, focusing more on hard, misclassified examples. The focal loss is given by formula \ref{formula:focal}:
\begin{equation}
\mathcal{L}_{\text{focal}} = -\alpha_t (1 - p_t)^{\gamma} \log(p_t)
\label{formula:focal}
\end{equation}
Where, \( \alpha_t \) is the balancing factor for class \( t \), \( p_t \) is the model's estimated probability for the correct class, and \( \gamma \) is the focusing parameter that adjusts the rate at which easy examples are down-weighted.
The hyperparameters \( \alpha_t \) and \( \gamma \) were determined using grid search as 0.35 and 4.0 respectively.\\
For Sub-task C, only decoder models were used during the Testing phase as BERT-based models massively underperformed in limited tests. An additional Gemma-2 27B model was fine-tuned for Sub-task B and C using Odds Ratio Preference Optimization (ORPO) \cite{hong2024orpomonolithicpreferenceoptimization} for better alignment.
All the fine-tuning of decoder-only models was carried out using Unsloth with Low-Rank Adaptation of Large Language Models (LoRA) \cite{hu2021loralowrankadaptationlarge}. The rank (\(r\)) and alpha (\(\alpha\)) values used were 16 for both.

\section{Results and Discussion}

\begin{table}[]
\centering
\resizebox{\columnwidth}{!}{
\begin{tabular}{|l|c|c|c|}
\hline
\textbf{Model}   & \textbf{F1}     & \textbf{Recall} & \textbf{Precision} \\ \hline
mBERT            & 0.9962          & 0.9962          & 0.9962             \\
mDistilBERT      & 0.9955          & 0.9957          & 0.9954             \\
XLM Roberta      & 0.9965          & 0.9966          & 0.9964             \\
MuRIL            & \textbf{0.9978} & \textbf{0.9978} & \textbf{0.9977}    \\
IndicBERT V2     & \textbf{0.9978} & \textbf{0.9978} & \textbf{0.9977}    \\
Llama 3.1 8B     & 0.9957          & 0.9957          & 0.9958             \\
Gemma-2 9B       & 0.9965          & 0.9965          & 0.9965             \\
Mistral Nemo 12B & 0.9962          & 0.9962          & 0.9961             \\ \hline
\end{tabular}
}
\caption{Performance metrics for Subtask A on the dev set}
\label{tab:taskAdev}
\end{table}

\begin{table}[]
\centering
\resizebox{\columnwidth}{!}{
\begin{tabular}{|l|l|c|}
\hline
\textbf{Model}        & \textbf{Description}        & \textbf{F1} \\ \hline
MuRIL        & Fine-tuned on train+dev set & 0.9968      \\
IndicBERT V2 & Fine-tuned on train+dev set & 0.9977      \\
Gemma-2 9B   & Fine-tuned on train+dev set & 0.9973      \\ \hline
Ensemble-1 & \begin{tabular}[c]{@{}l@{}}MuRIL's prediction as fallback \\ in case of no majority\end{tabular}       & 0.9979          \\
Ensemble-2 & \begin{tabular}[c]{@{}l@{}}IndicBERT V2's prediction as\\ fallback in case of no majority\end{tabular} & \textbf{0.9980} \\
Ensemble-3 & \begin{tabular}[c]{@{}l@{}}Gemma-2 9B's prediction as \\ fallback in case of no majority\end{tabular}  & 0.9979          \\ \hline
\end{tabular}
}
\caption{Performance metrics for Subtask A on the test set}
\label{tab:taskAtest}
\end{table}

\begin{table}[]
\centering
\resizebox{\columnwidth}{!}{
\begin{tabular}{|l|c|c|c|}
\hline
\textbf{Model}                 & \textbf{F1} & \textbf{Recall} & \textbf{Precision} \\ \hline
mBERT                          & 0.7142      & 0.7152          & 0.7133             \\
mDistilBERT                    & 0.6286      & 0.6093          & 0.6668             \\
XLM Roberta                    & 0.7182      & 0.7367          & 0.7037             \\
MuRIL                          & 0.6773      & \textbf{0.7741} & 0.6530             \\
IndicBERT V2                   & 0.7298      & 0.7215          & 0.7392             \\
Gemma-2 9B                     & 0.7094      & 0.6677          & \textbf{0.8051}    \\
Gemma-2 9B (Few-shot)          & \textbf{0.7412}  & 0.7019     & 0.7929             \\ \hline
\end{tabular}
}
\caption{Performance metrics for Subtask B on dev set}
\label{tab:taskBdev}
\end{table}

\begin{table}[]
\centering
\resizebox{\columnwidth}{!}{
\begin{tabular}{|l|c|c|c|}
\hline
\textbf{Model}              & \textbf{F1} & \textbf{Recall} & \textbf{Precision} \\ \hline
IndicBERT V2                & 0.7582      & 0.7732          & 0.7455             \\
Gemma-2 9B (Few-shot)       & 0.7588      & 0.7360          & 0.7895             \\
Gemma-2 27B Orpo            & 0.7494      & 0.7261          & 0.7814             \\ \hline
Ensemble                    & \textbf{0.7652}      & 0.7441          & 0.7925             \\ \hline
\end{tabular}
}
\caption{Performance metrics for Subtask B on test set}
\label{tab:taskBtest}
\end{table}

\begin{table}[]
\centering
\resizebox{\columnwidth}{!}{
\begin{tabular}{|l|c|c|c|}
\hline
\textbf{Model}        & \textbf{F1} & \textbf{Recall} & \textbf{Precision} \\ \hline
mDistilBERT  & 0.4173      & 0.4296          & 0.4560             \\
mBERT        & 0.4398      & 0.4567          & 0.4926             \\
XLM Roberta  & 0.5455      & 0.5765          & 0.5528             \\
IndicBERT V2 & 0.4639      & 0.4648          & 0.4643             \\
Gemma-2 9B   & \textbf{0.6937}      & \textbf{0.6691}          & \textbf{0.7520}             \\ \hline
\end{tabular}
}
\caption{Performance metrics for Subtask C on dev set}
\label{tab:taskCdev}
\end{table}

\begin{table*}[]
\centering
\resizebox{\textwidth}{!}{
\begin{tabular}{|l|l|c|c|c|}
\hline
\textbf{Model} &
  \textbf{Description} &
  \textbf{F1} &
  \textbf{Recall} &
  \textbf{Precision} \\ \hline
Gemma-2 9B &
  \begin{tabular}[c]{@{}l@{}}Fine-tuned on train+dev set with learning \\ rate 2e-4 and batch size of 4 for 2 epochs\end{tabular} &
  0.6213 &
  0.6084 &
  0.6734 \\
Gemma-2 9B &
  \begin{tabular}[c]{@{}l@{}}Fine-tuned on train+dev set with learning \\ rate 2e-4 and batch size of 2 for 2 epochs\end{tabular} &
  0.6503 &
  0.6371 &
  0.6982 \\
Gemma-2 27B &
  \begin{tabular}[c]{@{}l@{}}Fine-tuned on train+dev set using ORPO \\ with a batch size of 8 for 1 epoch\end{tabular} &
  \textbf{0.6804} &
  \textbf{0.6669} &
  \textbf{0.7183} \\ \hline
\end{tabular}%
}
\caption{Performance metrics for task C on the test set}
\label{tab:taskCtest}
\end{table*}

\subsection{Evaluation Phase}

During the Evaluation phase, various models were assessed across Sub-tasks A, B, and C using the dev set to identify the top-performing models for each task. For Sub-task A (Table \ref{tab:taskAdev}), the BERT-based models and decoder-only models, both delivered strong performances, with IndicBERT V2 and MuRIL emerging as the best models, each achieving an F1 score of \(0.9978\). They also had high recall and precision, indicating their robustness in effectively balancing sensitivity and specificity in task A classification. mBERT, XLM-Roberta, and larger generative models like Gemma-2 and Mistral Nemo also scored close to the top contenders, demonstrating that BERT-based and recent LLMs both possess considerable ability in text classification.
For Sub-task B (Table \ref{tab:taskBdev}), models' performance varied more significantly, reflecting the increased complexity compared to Sub-task A. Among the evaluated models, fine-tuned Gemma-2 9B with few-shot prompting yielded an F1 score of \(0.7412\). This shows Gemma-2’s effective adaptation in low-resource scenarios even with limited examples. IndicBERT V2 and XLM-Roberta also provided competitive results, with IndicBERT V2 achieving an F1 score of \(0.7298\), reinforcing its efficacy across both tasks. This marked Gemma-2 9B and IndicBERT V2 as the top choices to be further evaluated for Sub-task B during the Testing phase.
In Sub-task C (Table \ref{tab:taskCdev}), Gemma-2 9B demonstrated superior results with an F1 score of \(0.6937\). This outcome was significantly better than all other models, indicating Gemma-2’s robust performance for tasks with limited examples. XLM Roberta achieved the second-highest F1 score of \(0.5455\). The performance of other models shows the complexity of the task as except for Gemma-2, other models couldn't cross the F1 score of \(0.6\).

\subsection{Testing Phase}

For the testing phase, we retrained the top-selected models from the Evaluation phase by incorporating both the train and dev sets to create a more generalized model for final testing. For Sub-task A (Table \ref{tab:taskAtest}), ensemble techniques were applied to enhance accuracy further, leading to notable improvements in performance. Three ensembles were constructed, each with a different fallback model for cases without a majority prediction. Among these, Ensemble-2, which defaulted to IndicBERT V2's predictions when no majority was reached, yielded the highest F1 score of \(0.9980\). This ensemble strategy was instrumental in refining classification outcomes by leveraging the strengths of multiple models while relying on IndicBERT V2’s consistency as a fallback. As a result, Sub-task A saw an optimal performance boost, indicating the success of ensembling techniques in improving classification tasks with high base accuracy.
For Sub-task B (Table \ref{tab:taskBtest}), we employed a similar ensemble approach to maximize prediction performance. Ensemble results demonstrated improved robustness and balance across the metrics, culminating in an F1 score of \(0.7652\), with strong recall (\(0.7441\)) and precision (\(0.7925\)). For the ensemble, we employed an additional Gemma-2 27B trained using ORPO with the two models selected during the Evaluation phase. The overall gains from the ensemble approach for this task underscore its potential to improve tasks with more nuanced, challenging data patterns.
In Sub-task C (Table \ref{tab:taskCtest}), instead of using ensembling, we selected Gemma-2 27B ORPO as the optimal model for its strong performance during testing. This model achieved an F1 score of \(0.6804\), with balanced recall (\(0.6669\)) and precision (\(0.7183\)), showcasing its capability to handle more granular classification without the need for ensemble interventions. The decision to forego ensembling was based on the observation that Gemma-2 27B’s setup offered robust, reliable performance on its own, suggesting that, for some tasks, a single, finely-tuned model can sometimes match or exceed ensemble outcomes.

\section{Conclusion}
Our results demonstrate the importance of leveraging tailored approaches to tackle complex natural language understanding tasks across multiple languages in Devanagari script. By combining the multilingual strengths of the BERT-based models, focal loss for class sensitivity, and the generative power of Gemma-2, we achieved notable performance improvements across the subtasks. These findings highlight the value of adapting model architectures and training strategies to the nuances of each task, especially in handling multilingual contexts and imbalanced classes. This work lays a foundation for more refined, scalable hate speech detection systems for South Asian languages that can respond effectively to diverse and complex online discourse.

\section*{Limitations}
The datasets used for training and evaluation in hate speech and target detection are relatively small, which may impact the generalizability of the models in real-world applications. The challenges such as unbalanced datasets, difficulties in data collection, and issues with code-mixed languages, as noted in prior research \cite{parihar2021hate}, remain significant hurdles in the accurate detection of hate speech. Although techniques like focal loss and Odds Ratio Preference Optimization (ORPO) were applied to improve performance, the models still struggle with fine-grained distinctions in ambiguous hate speech contexts. Additionally, the decoder-only models were trained in 4-bit precision due to computational limitations, and they may perform better in full-precision mode. While these models performed well in most tasks, they are computationally intensive, requiring substantial resources for both fine-tuning and inference. On the other hand, BERT-based models performed well in Sub-tasks A and B, and with larger datasets, they may offer better performance for Sub-task C at a lower computational cost than decoder-only models.

\section*{Ethical Considerations}
When developing models for detecting hate speech and its targets, it’s important to address several ethical concerns. A major issue is the potential for bias in both the data and the model’s outputs. Since the datasets used in the development are limited and might not fully represent all social contexts, there’s a risk that the models could unintentionally reinforce biases or target specific groups unfairly. These models might also be used in ways that could cause harm, such as censoring or flagging content incorrectly without human oversight. Given the complex nuances of hate speech, it's crucial to avoid over-censorship, which may otherwise lead to the unjust targeting of certain communities or the stifling of legitimate free speech.
\bibliography{chipsal}
\appendix

\section{Appendix}
\label{sec:appendix}
\subsection{Confusion Matrix}
We provide the confusion matrix for all the models we tested below:
\subsubsection{Sub-task A: Language Detection}
\textbf{Evaluation Phase}
\begin{figure}[h]
    \centering
    \includegraphics[width=0.9\columnwidth]{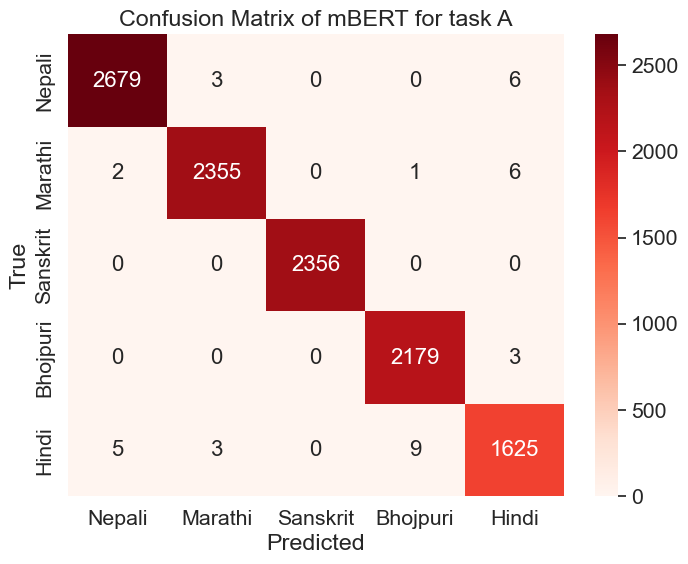} 
    \caption{mBERT's Confusion Matrix for Language Detection}
    \label{fig:confusion_matrix}
\end{figure}

\begin{figure}[H]
    \centering
    \includegraphics[width=0.9\columnwidth]{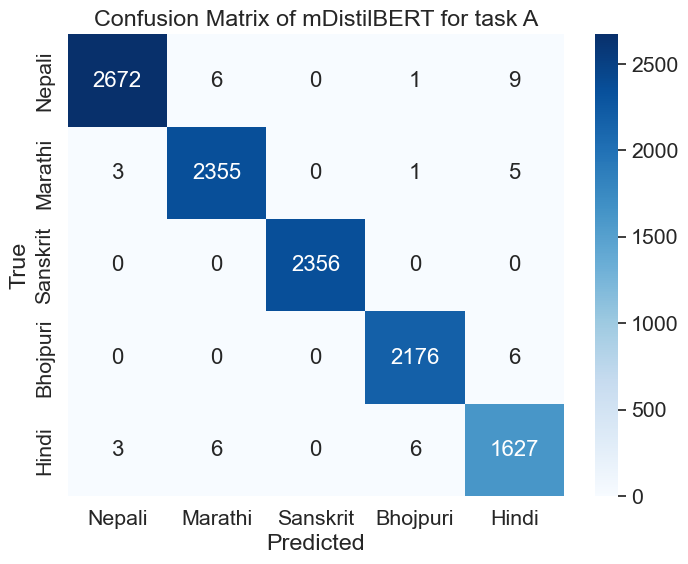} 
    \caption{mDistilBERT's Confusion Matrix for Language Detection}
    \label{fig:confusion_matrix}
\end{figure}

\begin{figure}[H]
    \centering
    \includegraphics[width=0.9\columnwidth]{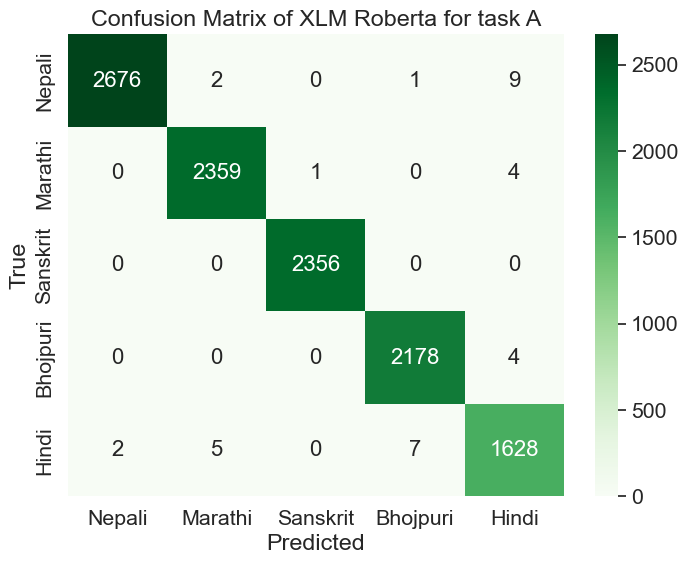} 
    \caption{XLM Roberta's Confusion Matrix for Language Detection}
    \label{fig:confusion_matrix}
\end{figure}

\begin{figure}[H]
    \centering
    \includegraphics[width=0.9\columnwidth]{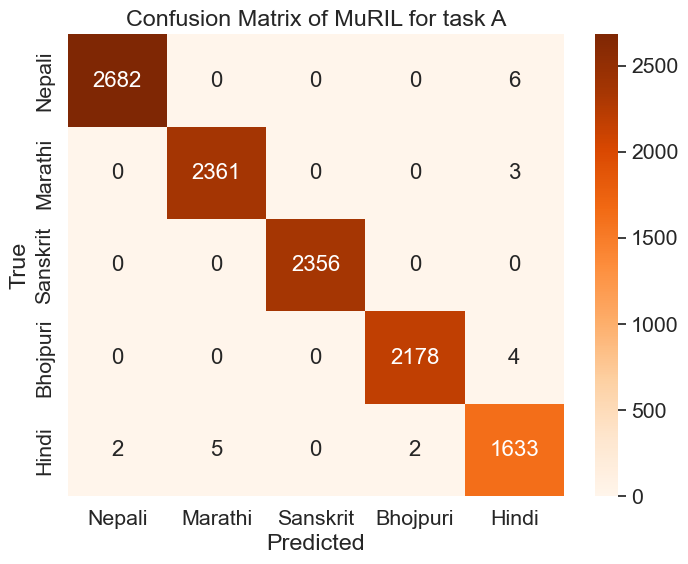} 
    \caption{MuRIL's Confusion Matrix for Language Detection}
    \label{fig:confusion_matrix}
\end{figure}

\begin{figure}[H]
    \centering
    \includegraphics[width=0.9\columnwidth]{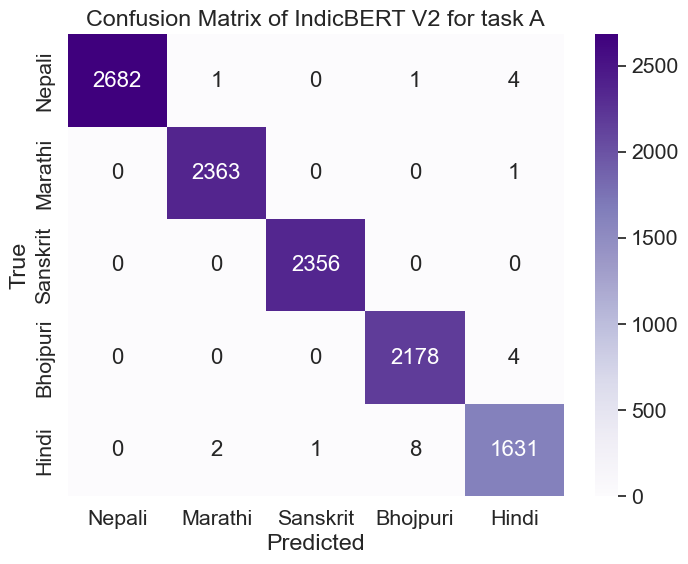} 
    \caption{IndicBERT V2's Confusion Matrix for Language Detection}
    \label{fig:confusion_matrix}
\end{figure}

\begin{figure}[H]
    \centering
    \includegraphics[width=0.9\columnwidth]{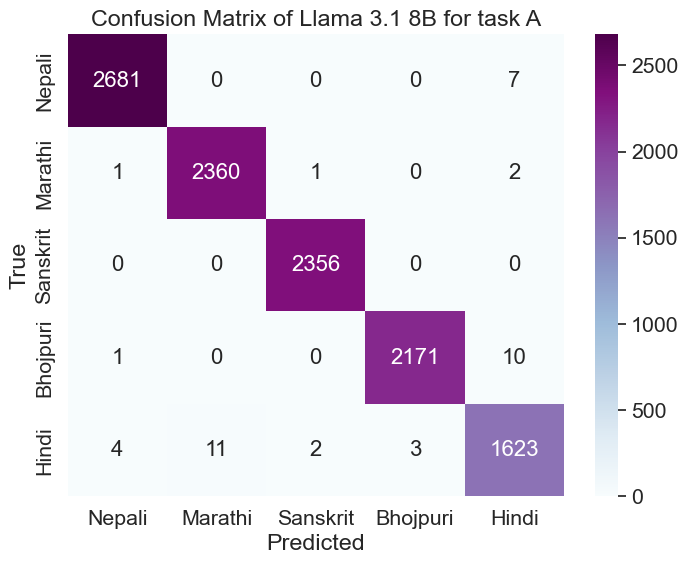} 
    \caption{Llama 3.1 8B's Confusion Matrix for Language Detection}
    \label{fig:confusion_matrix}
\end{figure}

\begin{figure}[H]
    \centering
    \includegraphics[width=0.9\columnwidth]{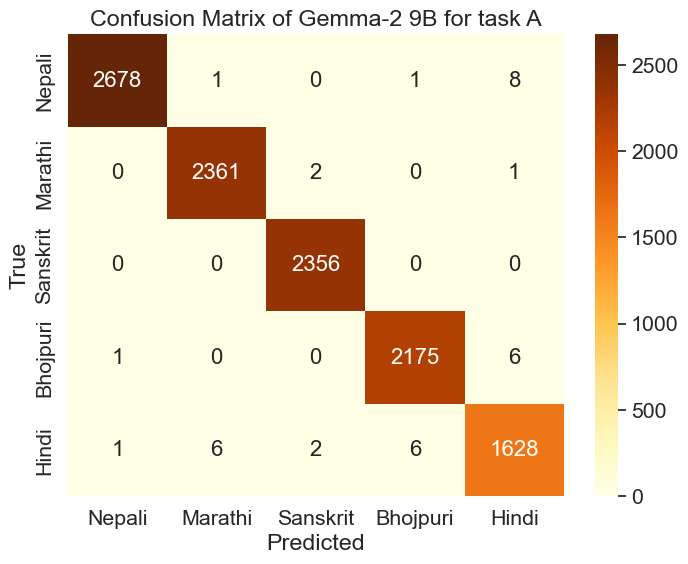} 
    \caption{Gemma-2 9B's Confusion Matrix for Language Detection}
    \label{fig:confusion_matrix}
\end{figure}

\begin{figure}[H]
    \centering
    \includegraphics[width=0.9\columnwidth]{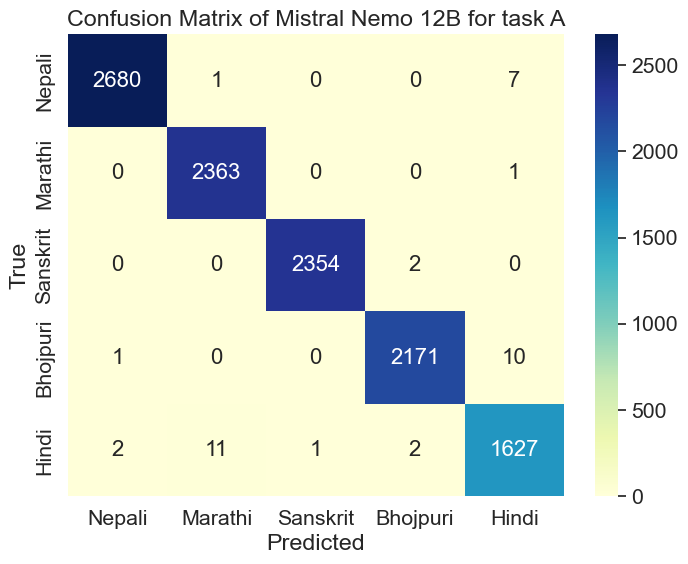} 
    \caption{Mistral Nemo's Confusion Matrix for Language Detection}
    \label{fig:confusion_matrix}
\end{figure}

\textbf{Testing Phase}
\begin{figure}[H]
    \centering
    \includegraphics[width=0.9\columnwidth]{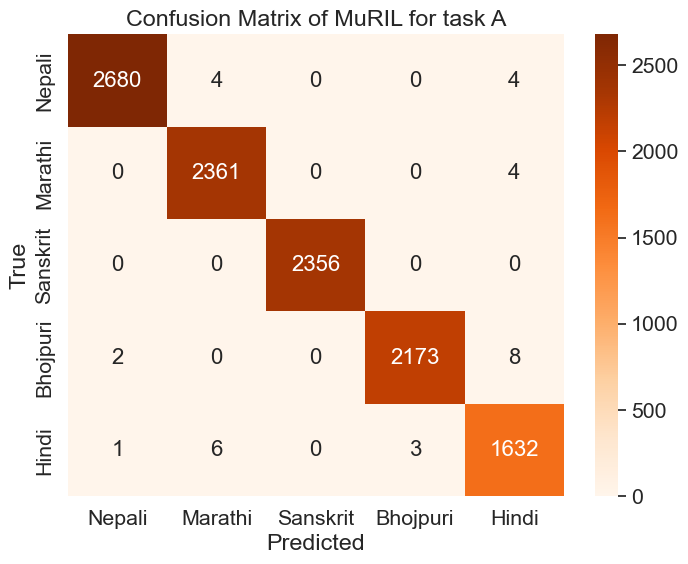} 
    \caption{MuRIL's Confusion Matrix for Language Detection}
    \label{fig:confusion_matrix}
\end{figure}

\begin{figure}[H]
    \centering
    \includegraphics[width=0.9\columnwidth]{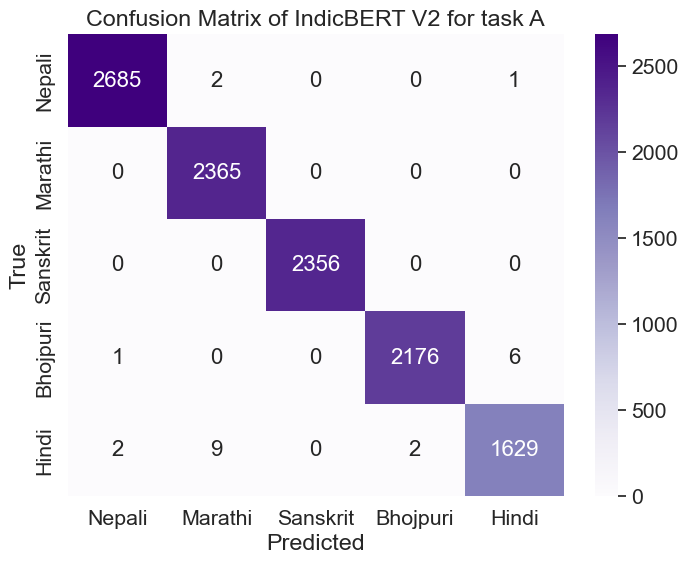} 
    \caption{IndicBERT V2's Confusion Matrix for Language Detection}
    \label{fig:confusion_matrix}
\end{figure}

\begin{figure}[H]
    \centering
    \includegraphics[width=0.9\columnwidth]{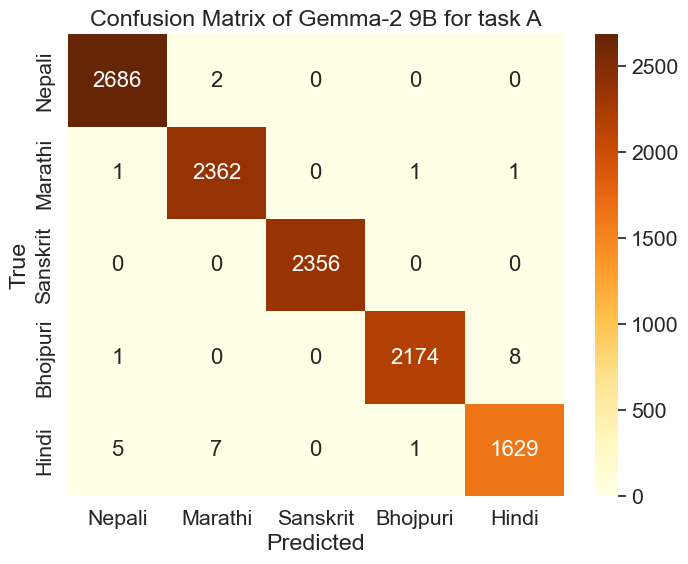} 
    \caption{Gemma-2 9B's Confusion Matrix for Language Detection}
    \label{fig:confusion_matrix}
\end{figure}

\begin{figure}[H]
    \centering
    \includegraphics[width=0.9\columnwidth]{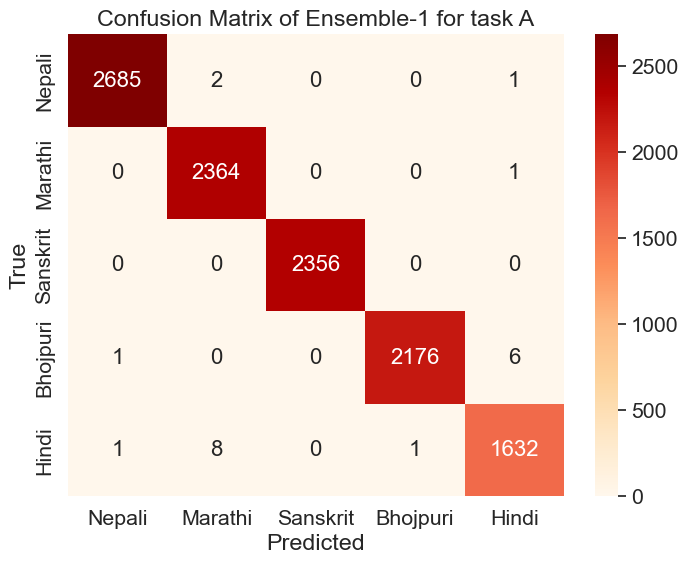} 
    \caption{Ensemble-1's Confusion Matrix for Language Detection}
    \label{fig:confusion_matrix}
\end{figure}

\begin{figure}[H]
    \centering
    \includegraphics[width=0.9\columnwidth]{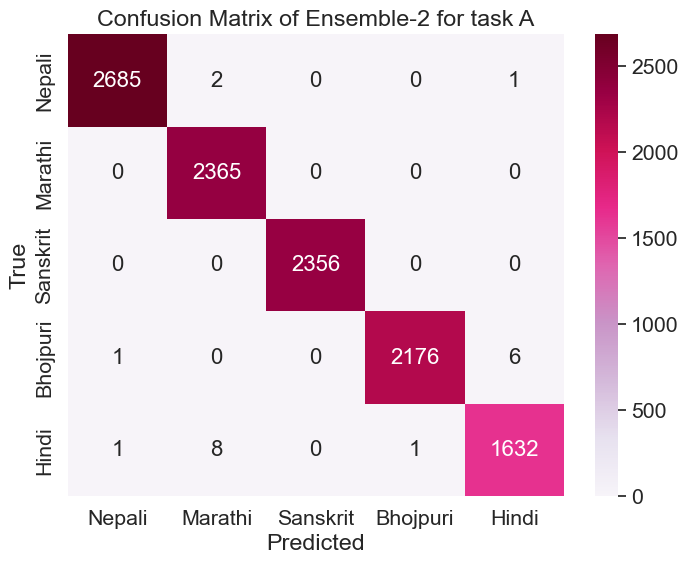} 
    \caption{Ensemble-2's Confusion Matrix for Language Detection}
    \label{fig:confusion_matrix}
\end{figure}

\begin{figure}[H]
    \centering
    \includegraphics[width=0.9\columnwidth]{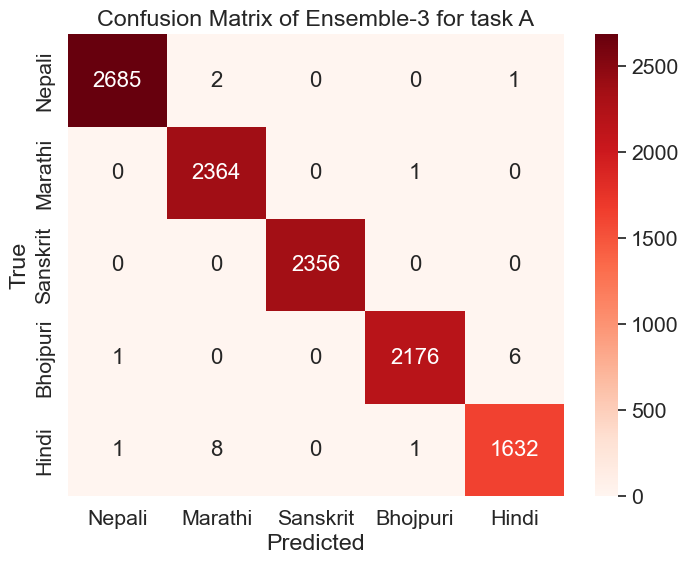} 
    \caption{Ensemble-3's Confusion Matrix for Language Detection}
    \label{fig:confusion_matrix}
\end{figure}

\subsubsection{Sub-task B: Hate Speech Detection}
\textbf{Evaluation Phase}
\begin{figure}[H]
    \centering
    \includegraphics[width=0.9\columnwidth]{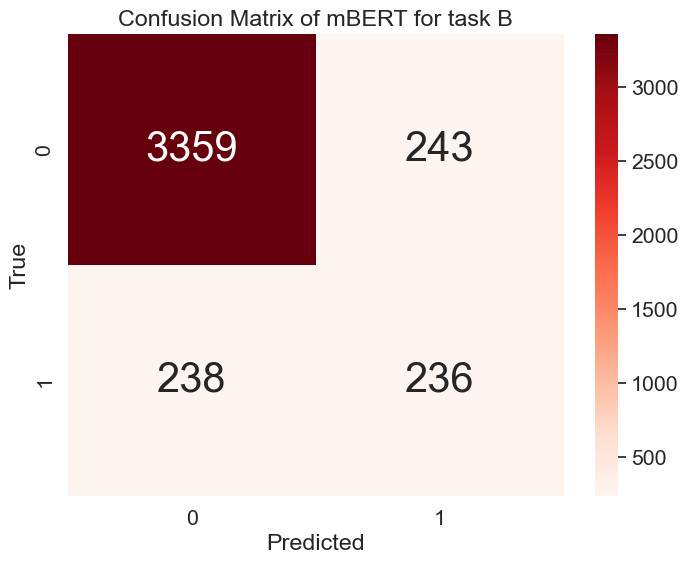} 
    \caption{mBERT's Confusion Matrix for Hate Speech Detection}
    \label{fig:confusion_matrix}
\end{figure}

\begin{figure}[H]
    \centering
    \includegraphics[width=0.9\columnwidth]{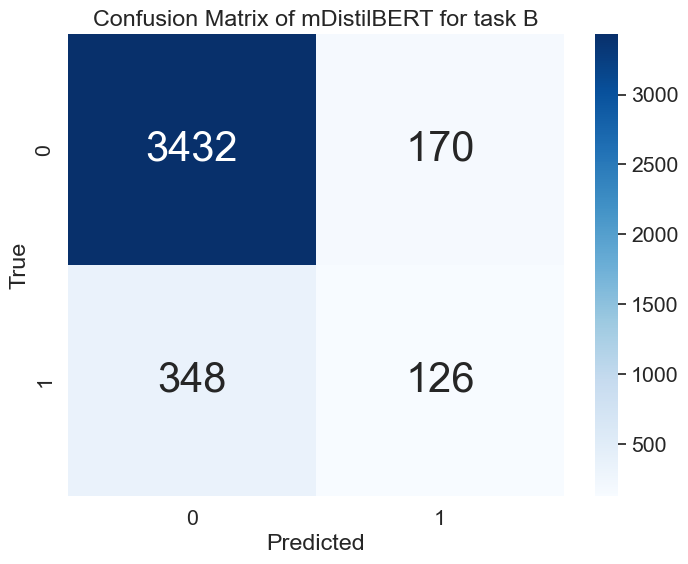} 
    \caption{mDistilBERT's Confusion Matrix for Hate Speech Detection}
    \label{fig:confusion_matrix}
\end{figure}

\begin{figure}[H]
    \centering
    \includegraphics[width=0.9\columnwidth]{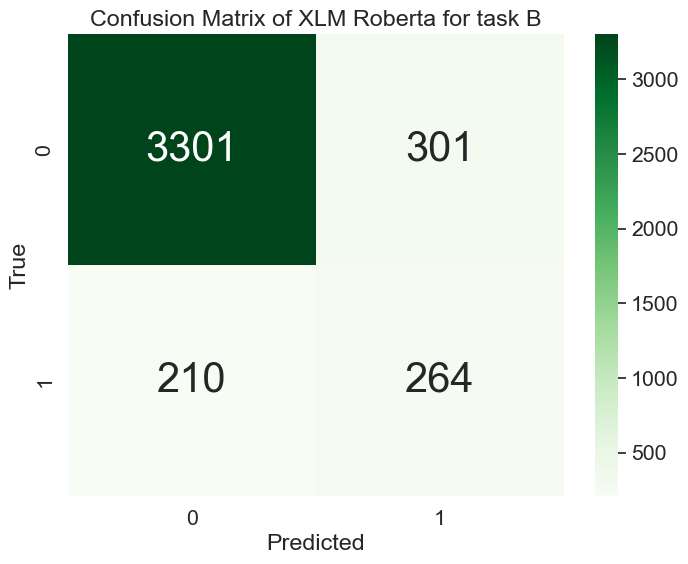} 
    \caption{XLM Roberta's Confusion Matrix for Hate Speech Detection}
    \label{fig:confusion_matrix}
\end{figure}

\begin{figure}[H]
    \centering
    \includegraphics[width=0.9\columnwidth]{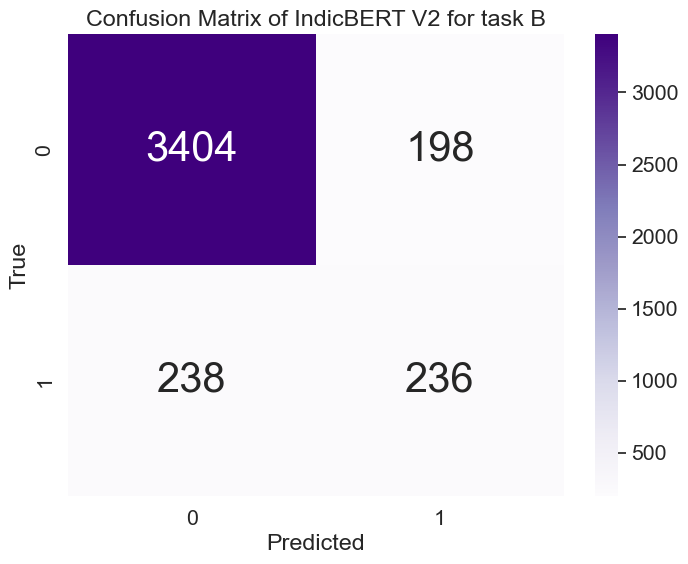} 
    \caption{IndicBERT V2's Confusion Matrix for Hate Speech Detection}
    \label{fig:confusion_matrix}
\end{figure}

\begin{figure}[H]
    \centering
    \includegraphics[width=0.9\columnwidth]{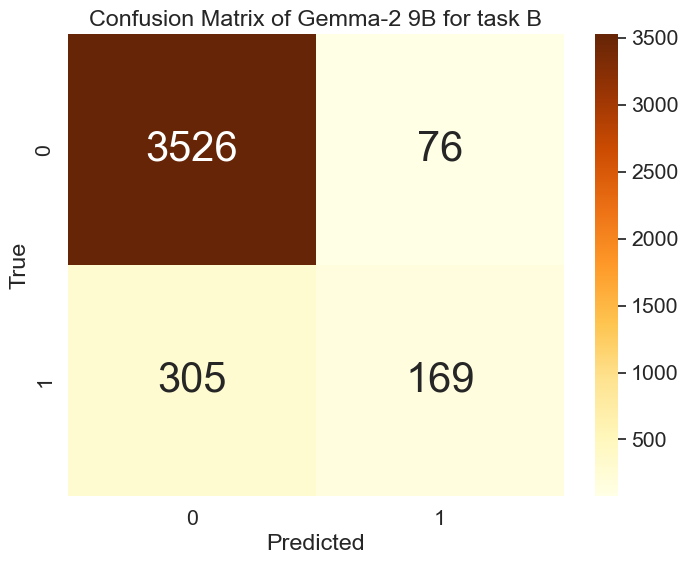} 
    \caption{Gemma-2 9B's Confusion Matrix for Hate Speech Detection}
    \label{fig:confusion_matrix}
\end{figure}

\begin{figure}[H]
    \centering
    \includegraphics[width=0.9\columnwidth]{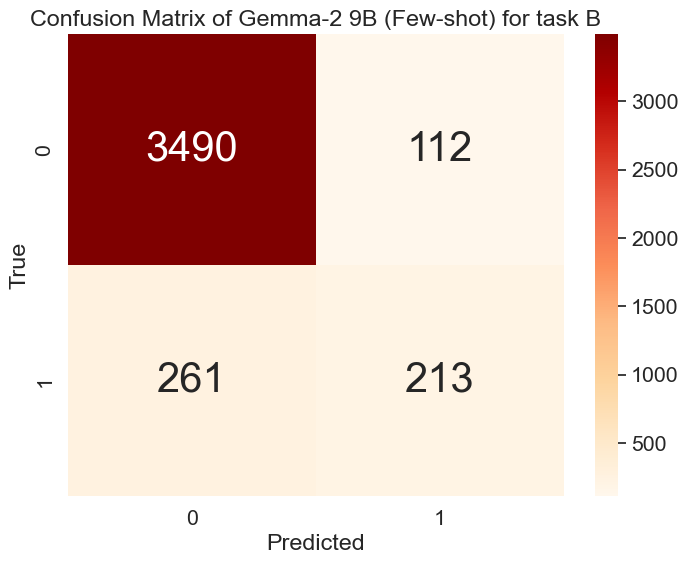} 
    \caption{Gemma-2 9B (Few-shot)'s Confusion Matrix for Hate Speech Detection}
    \label{fig:confusion_matrix}
\end{figure}

\textbf{Testing Phase}

\begin{figure}[H]
    \centering
    \includegraphics[width=0.9\columnwidth]{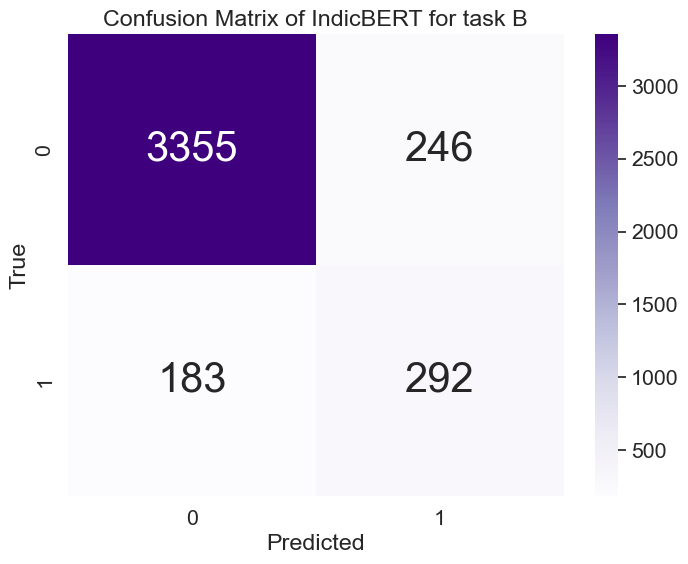} 
    \caption{IndicBERT V2's Confusion Matrix for Hate Speech Detection}
    \label{fig:confusion_matrix}
\end{figure}

\begin{figure}[H]
    \centering
    \includegraphics[width=0.9\columnwidth]{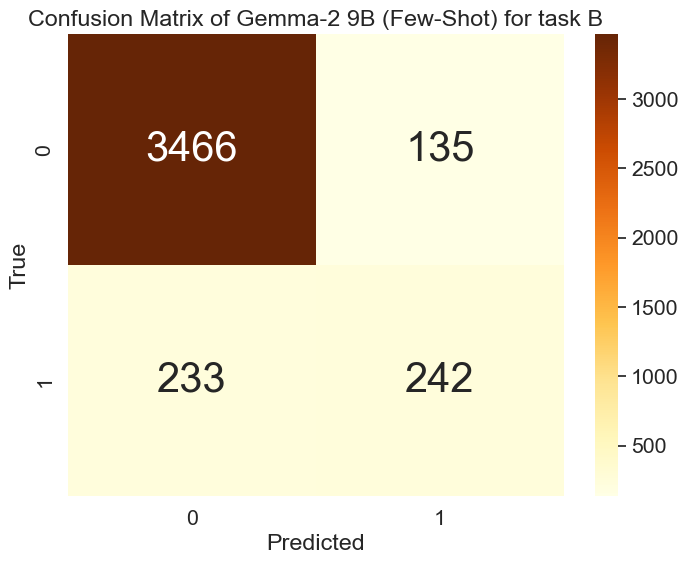} 
    \caption{Gemma-2 9B (Few-shot)'s Confusion Matrix for Hate Speech Detection}
    \label{fig:confusion_matrix}
\end{figure}

\begin{figure}[H]
    \centering
    \includegraphics[width=0.9\columnwidth]{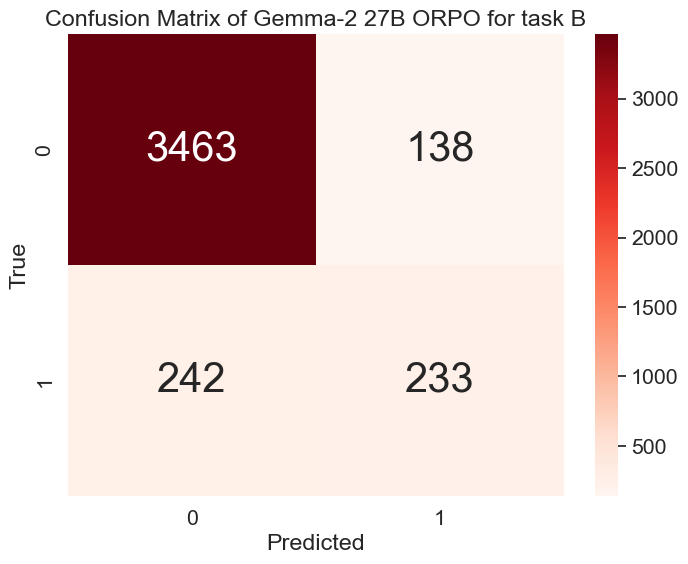} 
    \caption{Gemma-2 27B ORPO's Confusion Matrix for Hate Speech Detection}
    \label{fig:confusion_matrix}
\end{figure}

\begin{figure}[H]
    \centering
    \includegraphics[width=0.9\columnwidth]{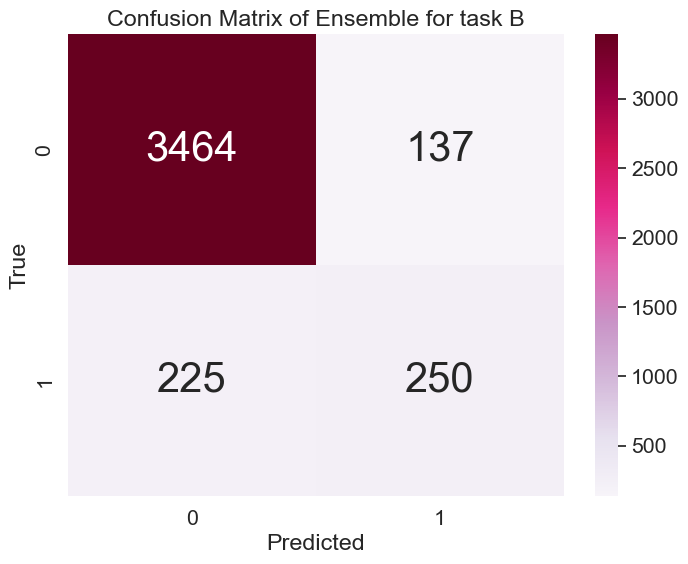} 
    \caption{Ensemble's Confusion Matrix for Hate Speech Detection}
    \label{fig:confusion_matrix}
\end{figure}

\subsubsection{Sub-task C: Hate Speech Target Detection}
\textbf{Evaluation Phase}
\begin{figure}[H]
    \centering
    \includegraphics[width=0.9\columnwidth]{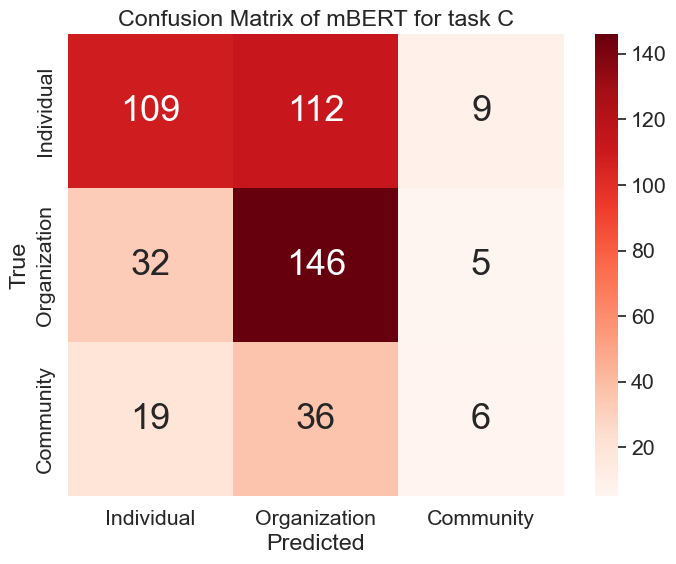} 
    \caption{mBERT's Confusion Matrix for Hate Speech Target Detection}
    \label{fig:confusion_matrix}
\end{figure}

\begin{figure}[H]
    \centering
    \includegraphics[width=0.9\columnwidth]{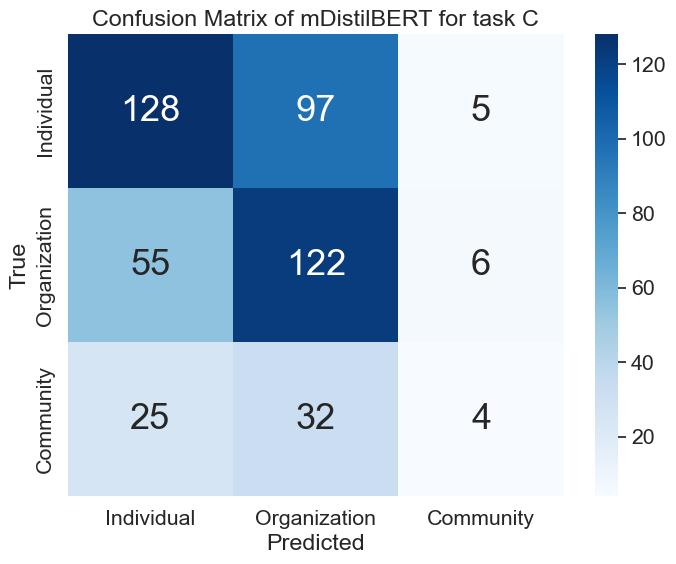} 
    \caption{mDistilBERT's Confusion Matrix for Hate Speech Target Detection}
    \label{fig:confusion_matrix}
\end{figure}

\begin{figure}[H]
    \centering
    \includegraphics[width=0.9\columnwidth]{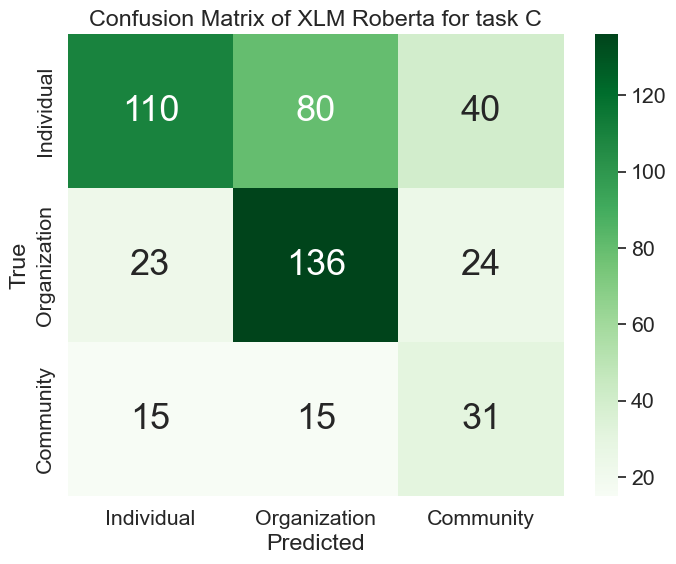} 
    \caption{XLM Roberta's Confusion Matrix for Hate Speech Target Detection}
    \label{fig:confusion_matrix}
\end{figure}

\begin{figure}[H]
    \centering
    \includegraphics[width=0.9\columnwidth]{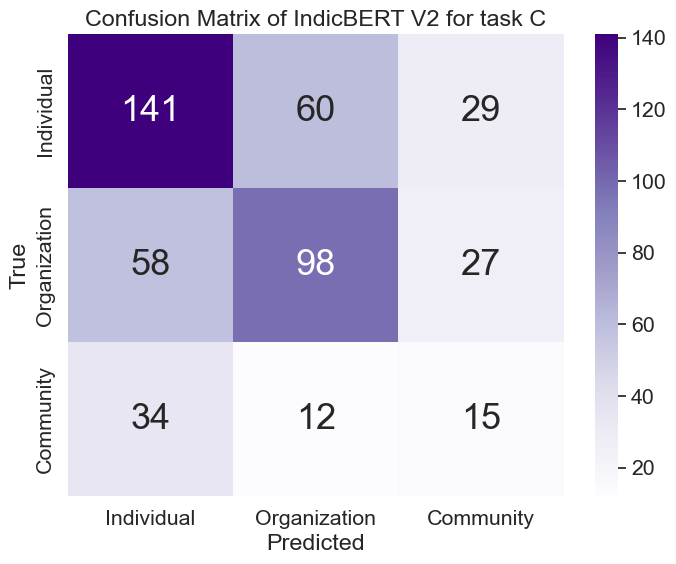} 
    \caption{IndicBERT V2's Confusion Matrix for Hate Speech Target Detection}
    \label{fig:confusion_matrix}
\end{figure}

\begin{figure}[H]
    \centering
    \includegraphics[width=0.9\columnwidth]{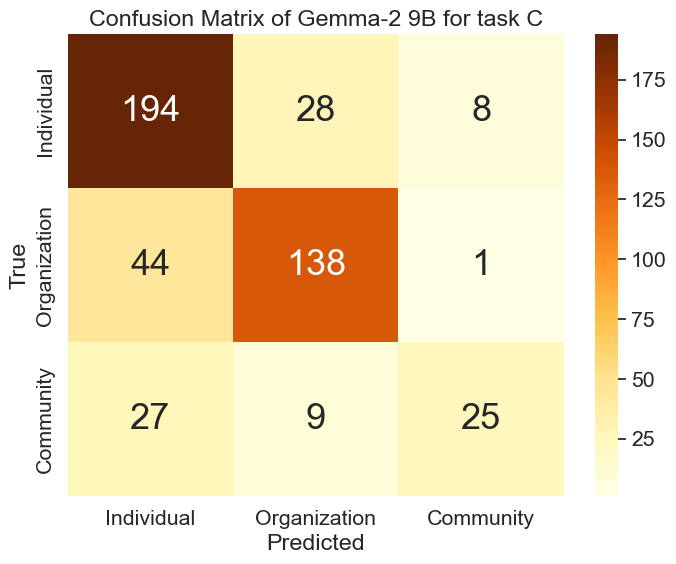} 
    \caption{Gemma-2 9B's Confusion Matrix for Hate Speech Target Detection}
    \label{fig:confusion_matrix}
\end{figure}

\textbf{Testing Phase}

\begin{figure}[H]
    \centering
    \includegraphics[width=0.9\columnwidth]{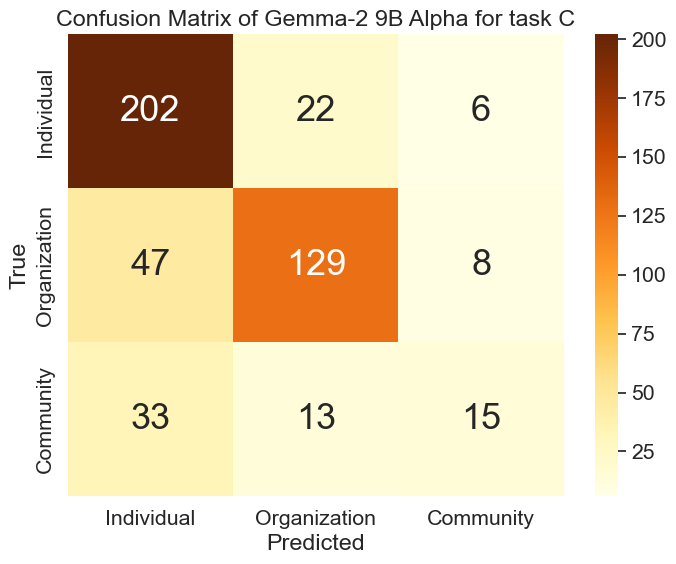} 
    \caption{Gemma-2 9B Alpha's Confusion Matrix for Hate Speech Target Detection}
    \label{fig:confusion_matrix}
\end{figure}

\begin{figure}[H]
    \centering
    \includegraphics[width=0.9\columnwidth]{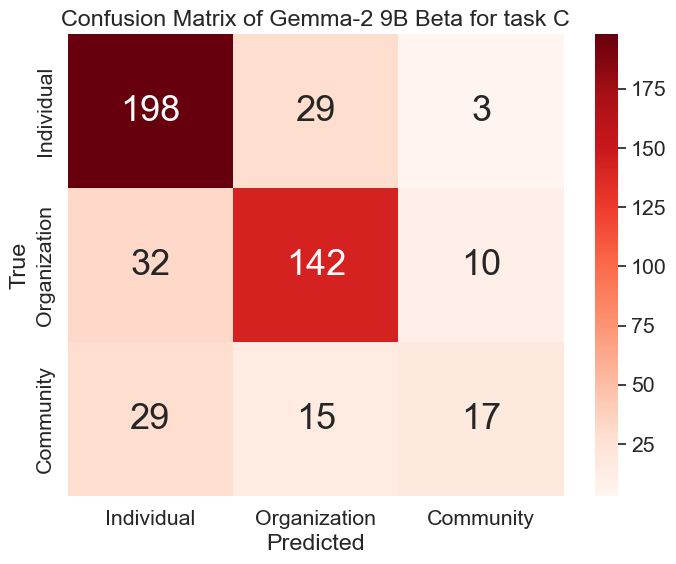} 
    \caption{Gemma-2 9B Beta's Confusion Matrix for Hate Speech Target Detection}
    \label{fig:confusion_matrix}
\end{figure}

\begin{figure}[H]
    \centering
    \includegraphics[width=0.9\columnwidth]{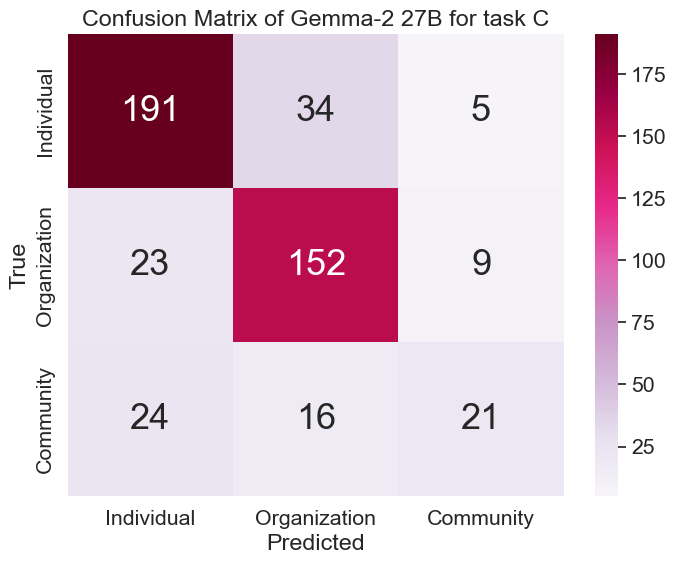} 
    \caption{Gemma-2 27B's Confusion Matrix for Hate Speech Target Detection}
    \label{fig:confusion_matrix}
\end{figure}

\subsection{System Replication}
We provide the details of hyperparameters used in training for replicating the process in Table \ref{tab:hyperd} and \ref{tab:hyperb}.
\begin{table}[H]
\centering
\begin{tabular}{|l|c|c|c|}
\hline
\textbf{Hyperparameter}          & \textbf{Value} \\ \hline
Learning rate                    & 2e-4           \\
Learning rate scheduler          & linear         \\
Weight decay                     & 0.01           \\
LoRA rank                        & 16             \\
LoRA alpha                       & 16             \\
LoRA dropout                     & 0              \\ \hline
\multicolumn{2}{l}{\textbf{Language Detection}} \\ \hline
Max length (tokens)              & 2048           \\
Batch size                       & 9              \\
Gradient accumulation            & 3              \\
Warmup steps                     & 5              \\
Num of epochs                    & 1              \\ \hline
\multicolumn{2}{l}{\textbf{Hate Speech Detection}} \\ \hline
Max length (tokens)              & 1024           \\
Batch size                       & 16             \\
Gradient accumulation            & 1              \\
Warmup steps                     & 10             \\
Num of epochs                    & 2-4            \\ \hline
\multicolumn{2}{l}{\textbf{Hate Speech Target Detection}} \\ \hline
Max length (tokens)              & 1024           \\
Batch size                       & 2-4            \\
Gradient accumulation            & 1              \\
Warmup steps                     & 0              \\
Num of epochs                    & 2              \\ \hline
\end{tabular}
\caption{Hyperparameters' values for decoder-only models across tasks}
\label{tab:hyperd}
\end{table}

\begin{table}[H]
\centering
\begin{tabular}{|l|c|}
\hline
\textbf{Hyperparameter}               & \textbf{Values} \\ \hline
\textbf{Max length of input sequence} & 64              \\
\textbf{Batch size}                   & 512             \\
\textbf{Num of workers}               & 2               \\
\textbf{Num of epochs}                & 5               \\
\textbf{Learning rate}                & 4e-5            \\
\textbf{Learning rate scheduler}      & linear          \\
\textbf{Focal loss Alpha}      & 0.35          \\
\textbf{Focal loss Gamma}      & 4.0          \\ \hline
\end{tabular}
\caption{Hyperparameters' values for BERT-based models}
\label{tab:hyperb}
\end{table}

Table \ref{tab:hyperd} presents the hyperparameters for decoder-only models across tasks, with core values, such as learning rate, weight decay,  and LoRA values shared across tasks. Task-specific parameters like maximum token length, batch size, gradient accumulation, warmup steps, and epochs were experimented with to meet the requirements of each task. For hyperparameters not listed, default values were used for each model.

\subsection{Prompts}
The prompts used for decoder-only models are provided below:

\lstset{
    basicstyle=\small\ttfamily,  
    breaklines=true,             
    breakatwhitespace=true,      
    frame=single,                
    columns=fullflexible         
}

\subsubsection{Task A: Language Detection}
\begin{lstlisting}
Task: You are an expert linguist specializing in Devanagari script languages. Your task is to identify the language of the given text.

### Instruction:
Analyze the following Devanagari script text and determine its language. Choose the correct language code from these options:
0: Nepali
1: Marathi
2: Sanskrit
3: Bhojpuri
4: Hindi

### Input:
Text: {text}

### Response:
The language code for the given text is: {label}
\end{lstlisting}

\subsubsection{Task B: Hate Speech Detection}
\begin{lstlisting}
Task: You are fluent in Nepali and Hindi languages. Your task is to classify if the given input text contains hate speech or not.

### Instruction:
The goal of this subtask is to identify the targets of hate speech in a given text. Choose the correct category from these options:
1: Hate
0: Non-Hate

### Examples:
Input: {example_text1} 
Response: {example_text1_label} 

Input: {example_text2} 
Response: {example_text2_label} 

Input: {example_text3} 
Response: {example_text3_label} 

Input: {example_text4} 
Response: {example_text4_label} 

Input: {example_text5} 
Response: {example_text5_label} 

### Input:
{text}

### Response:
{label}
\end{lstlisting}

\subsubsection{Task C: Hate Speech Target Detection}
\begin{lstlisting}
You are an expert linguist specializing in detecting hate speech targets in Devanagari-script tweets. Your task is to classify the target of hate speech.

### Instruction:
Analyze the given tweet in Devanagari script and determine who the hate speech is targeting.

Step 1: First, decide if the target is an individual or a group.
Step 2 (if group): If it's a group, further classify it as either an organization or a community.

Classify the final label according to these categories:
0. Individual: A specific person or a small set of identifiable individuals
1. Organization: A formal entity, institution, or company
2. Community: A broader group based on ethnicity, religion, gender, or other shared characteristics

### Input:
{}

### Response:
{}
\end{lstlisting}
\end{document}